\documentclass[conference]{IEEEtran}
\IEEEoverridecommandlockouts
\usepackage[switch]{lineno}
\usepackage{multicol}
\usepackage{cite}
\usepackage{amsmath,amssymb,amsfonts}
\usepackage{algorithmic}
\usepackage{graphicx}
\usepackage{textcomp}
\usepackage{xcolor}
\usepackage{url}
\usepackage{booktabs}
\usepackage{multirow}
\usepackage[colorlinks=true, citecolor=blue]{hyperref}
\def\BibTeX{{\rm B\kern-.05em{\sc i\kern-.025em b}\kern-.08em
    T\kern-.1667em\lower.7ex\hbox{E}\kern-.125emX}}

\usepackage{latexsym}
\usepackage{amsthm}
\usepackage{enumitem}
\usepackage[table]{xcolor}
\usepackage[ruled,linesnumbered]{algorithm2e}
\usepackage[inkscapelatex=false]{svg}
\usepackage{color}
\usepackage{anyfontsize}

\definecolor{bestblue}{HTML}{A5D1F8}    
\definecolor{secondblue}{HTML}{D9E8FB} 

\definecolor{darkred}{RGB}{139, 0, 0}
\definecolor{darkblue}{RGB}{0, 0, 139}

\begin{document}

\title{DE3S: Dual-Enhanced Soft-Sparse Shape Learning for Medical Early Time Series Classification\vspace{-1em}}

\author{
\\
\\
\IEEEauthorblockN{Tao Xie\textsuperscript{${1,\dagger}$}, Zexi Tan\textsuperscript{${1,\dagger}$}, Haoyi Xiao$^{1}$, Binbin Sun$^{2}$, Yiqun Zhang$^{1,3,*}$\thanks{$\dagger$ Co-first author\ \ \ $*$ Corresponding author}}

\IEEEauthorblockA{\textit{$^1$Guangdong University of Technology}\ \ \textit{$^2$Shenzhen Maternity and Child Healthcare Hospital}\ \ \textit{$^3$Hong Kong Baptist University}\\
\{3123001085, 3123004194, 3124004599\}@mail2.gdut.edu.cn, sunbin13530280620@126.com, yqzhang@gdut.edu.cn
}

\thanks{This work was supported in part by the National Natural Science Foundation of China (NSFC) under Grant 62476063, the Natural Science Foundation of Guangdong Province under Grant 2025A1515011293, the General Projects of Shenzhen Science and Technology Program under Grant JCYJ20240813115124032, and the Shenzhen Maternity and Child Healthcare Hospital under Grant FYA2022018.}
}


\maketitle

\begin{abstract}
Early Time Series Classification (ETSC) is critical in time-sensitive medical applications such as sepsis, yet it presents an inherent trade-off between accuracy and earliness. This trade-off arises from two core challenges: 1) models should effectively model inherently weak and noisy early-stage snippets, and 2) they should resolve the complex, dual requirement of simultaneously capturing local, subject-specific variations and overarching global temporal patterns. Existing methods struggle to overcome these underlying challenges, often forcing a severe compromise: sacrificing accuracy to achieve earliness, or vice-versa. We propose \textbf{DE3S}, a \textbf{D}ual-\textbf{E}nhanced \textbf{S}oft-\textbf{S}parse \textbf{S}equence Learning framework, which systematically solves these challenges. A dual enhancement mechanism is proposed to enhance the modeling of weak, early signals. Then, an attention-based patch  module is introduced to preserve discriminative information while reducing noise and complexity. A dual-path fusion architecture is designed, using a sparse mixture of experts to model local, subject-specific variations. A multi-scale inception module is also employed to capture global dependencies. Experiments on six real-world medical datasets show the competitive performance of DE3S, particularly in early prediction windows. Ablation studies confirm the effectiveness of each component in addressing its targeted challenge. The source code is available \href{https://github.com/kuxit/DE3S}{\textbf{here}}. 
\end{abstract}

\begin{IEEEkeywords}
Time Series Classification; Early Medical Prediction; Mixture of Experts; Sepsis Prediction.
\end{IEEEkeywords}

\section{Introduction}
Early Time Series Classification (ETSC) in medical applications is critical for patient outcomes, particularly in life-threatening conditions such as sepsis~\cite{Gupta2020, Tang2024, Ibarz2024}. The ability to identify disease patterns from limited temporal observations can significantly improve treatment outcomes and reduce mortality rates~\cite{Giannoula2018TemporalPatterns}. Medical time series classification presents unique challenges that distinguish it from conventional time series analysis: severe class imbalance, complex temporal patterns that vary across individuals, and the critical need for accurate early diagnosis with limited temporal observations~\cite{ismail2019deep}.

ETSC aims to classify time series before observing full data, which is vital in time-sensitive applications like sepsis diagnosis in ICUs, providing doctors more opportunities to save lives~\cite{shi2024time}. However, ETSC involves two conflicting goals of accuracy and earliness. Most existing methods balance these by trading one against the other, but a powerful early classifier should make highly accurate predictions at any moment~\cite{liu2024moirai}. The main obstacle is that key classification features are not readily apparent in the early stages, leading to overlapping distributions across time stages and difficulty in differentiation~\cite{zhang2023learning}, which is exacerbated by class imbalance and subject-specific variations.

The fundamental challenge in early medical time series classification lies in the dual requirements to capture both \textbf{local discriminative patterns} and \textbf{global temporal dependencies} while maintaining robustness to class imbalance and individual differences~\cite{Wang2024_survey, Stonko2024, Lin2024}. Local patterns, such as specific physiological signatures that emerge early in disease progression, are crucial for early detection. In the area of sepsis prediction, subtle changes in vital signs like heart rate variability and blood pressure patterns can serve as early indicators~\cite{Yin2024}. However, these patterns must be understood within the broader temporal context to avoid false positives and capture the full disease trajectory. The interplay between local and global information is particularly critical in medical applications where both fine-grained physiological changes and overall patient trajectory are essential for accurate diagnosis~\cite{Zhang2024_LogoRA}.

Traditional approaches have primarily focused on either local pattern recognition through shapelet-based methods or global temporal modeling through recurrent neural networks. Shapelet-based methods excel at identifying discriminative local patterns but struggle with capturing global temporal context \cite{ye2009time, Li2021ShapeNet}, while attention-based models~\cite{ye2024mad, ye2025mssnet} can capture long-range dependencies but may miss critical local patterns that are essential for early detection. This limitation becomes particularly problematic in medical applications where both local and global information are crucial for accurate diagnosis. Furthermore, existing methods often fail to address the computational complexity of shapelets discovery and the challenge of handling subject-specific variations in medical data~\cite{tan2025}.

\begin{figure*}[!t]
\centering
\includegraphics[width=\textwidth]{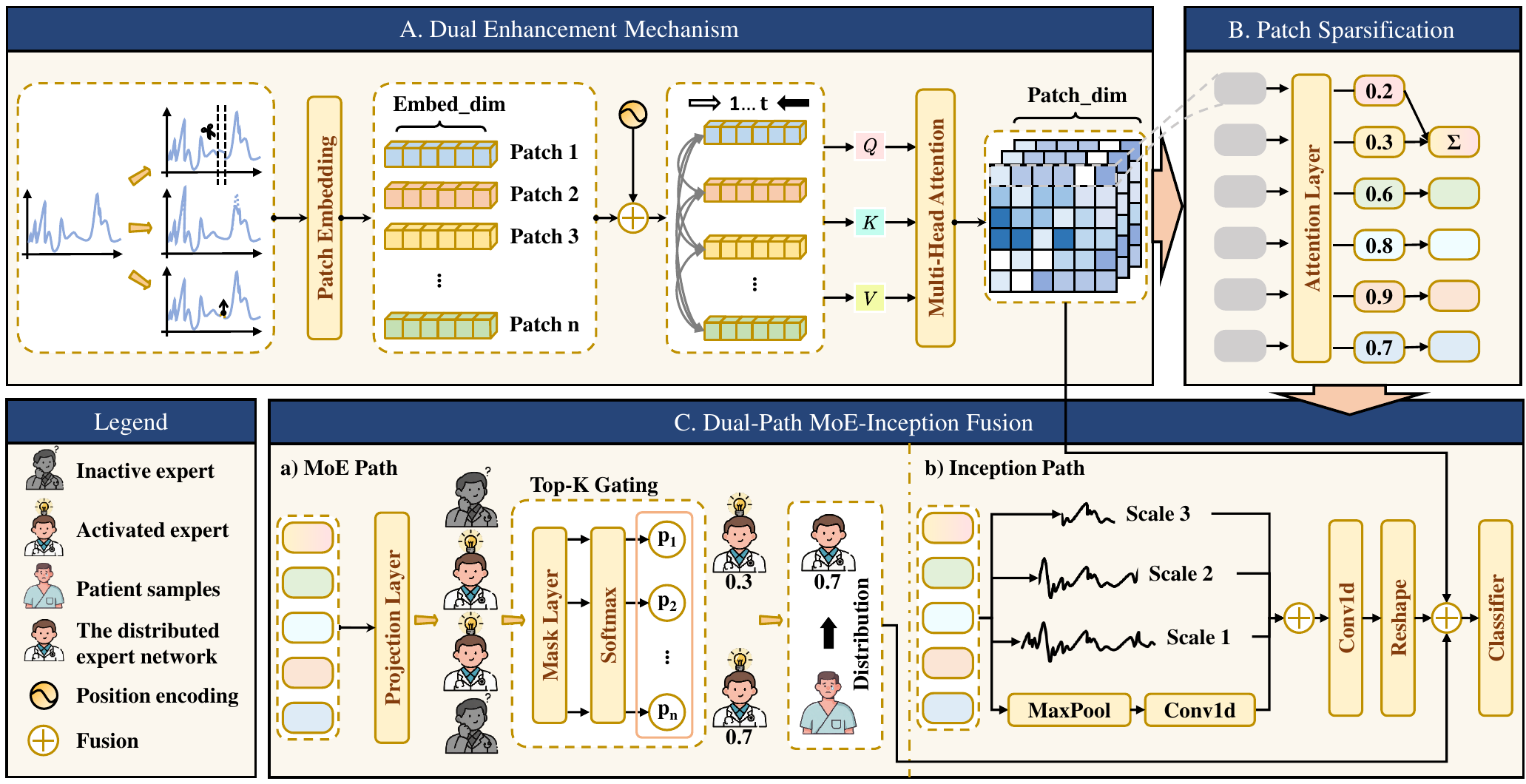}
\caption{\textbf{Overall framework of DE3S.} The architecture implements our core methodology: (1) A \textbf{Dual Enhancement Mechanism} first applies data augmentation and then uses patch embedding and multi-head attention for model-level enhancement. (2) An \textbf{Attention-based Patch Sparsification} module selects and aggregates patch tokens. (3) A \textbf{Dual-Path Fusion} module processes the tokens for final classification.}
\label{fig:framework}
\end{figure*}

To address these challenges, this paper proposes \textbf{DE3S}: a novel \textbf{Dual-Enhanced Soft-Sparse-Shape Learning} method. 
This method is designed to be robust against weak early-stage signals and subject-specific variations, which are common challenges in medical time series analysis. Instead of relying on a single enhancement method, it integrates a comprehensive dual-enhancement mechanism, an attention-based patch  mechanism, and a dual-path MoE-Inception fusion architecture. This holistic approach systematically strengthens weak signals, efficiently identifies discriminative patterns (patches), and captures both local and global temporal dependencies to improve both the accuracy and earliness of ETSC in critical medical applications. The main three contributions of this work are summarized as follows:

\begin{itemize}
    \item \textbf{Dual Enhancement Mechanism}: We develop a dual-enhancement mechanism that synergistically strengthens data- and model-level features, amplifying weak early signals while preserving discriminative characteristics. 

    \item \textbf{Attention-based Patch Sparsification}: We propose a soft  mechanism leveraging attention scores to dynamically aggregate less important patch tokens while preserving discriminative ones, thereby enhancing the accuracy and efficiency of early-stage classification.

    \item \textbf{Dual-Path MoE-Inception Fusion}: We propose a Dual-Path MoE-Inception Fusion that synergizes local learning (MoE) within individual patches with global pattern capture (Inception) across them, ensuring comprehensive temporal modeling of both local and global dependencies.
\end{itemize}

\section{Proposed Method}
This section details the \textbf{D}ual-\textbf{E}nhanced Soft-\textbf{S}parse \textbf{S}equence \textbf{L}earning (DE3S) framework. The methodology comprises three core components. First, the \textbf{Dual Enhancement Mechanism} is designed to amplify weak signals at both the data and model levels. Second, an \textbf{Attention-based Patch Sparsification} strategy is employed for efficient and focused representation learning. Finally, the \textbf{Dual-Path MoE-Inception Fusion} architecture captures both subject-specific and global temporal patterns.

\subsection{Overall Architecture}
\label{sec:overview}

As illustrated in Fig.~\ref{fig:framework}, the proposed architecture systematically addresses these challenges.
(1) The \textbf{Dual Enhancement Mechanism} first strengthens the input data and then creates context-aware patch tokens.
(2) The \textbf{Attention-based Patch Sparsification} module then selects and aggregates these tokens for efficiency.
(3) The resulting tokens are fed into the \textbf{Dual-Path MoE-Inception Fusion} module for comprehensive feature extraction. 
Finally, an attention-weighted classification head produce the final prediction.

\subsection{Dual Enhancement Mechanism}
\label{sec:enhancement}
The first component of our framework is a synergistic enhancement mechanism that addresses the weak signal problem from two perspectives: strengthening the raw input data and enriching the model-level representations.

\textbf{Data-Level Enhancement.}
To simulate real-world clinical variations and improve model robustness, the proposed method first apply a series of augmentations to the raw time series $\mathbf{X}$ during training. This includes: (1) \textit{Temporal cropping} to mimic different observation windows, (2) \textit{Amplitude scaling} to account for sensor calibration variance, and (3) \textit{Gaussian noise injection} to simulate measurement artifacts. This operation can be described by the following formula: 
\begin{equation}
\mathbf{X}_{\text{aug}} = \text{Noise}(\text{Scale}(\text{Crop}(\mathbf{X}))) \text{.}
\label{eq:aug}
\end{equation}

\textbf{Model-Level Enhancement.}
Following data-level enhancement, model-level enhancement is performed. This process comprises two steps: patch embedding and contextual enhancement. The initial step, patch embedding, tokenizes the time series into a sequence of patch representations. This tokenization is achieved using a 1D convolution as an efficient patch embedder, which slices the series $\mathbf{X}_{\text{aug}}$ into overlapping windows and embeds them into a $D$-dimensional space. Positional encoding $\mathbf{P}$ is added to retain temporal order. This yields an initial patch token sequence $\mathbf{S}$:
\begin{equation}
\mathbf{S} = \text{Conv1D}(\mathbf{X}_{\text{aug}}) + \mathbf{P} \text{,}
\label{eq:embed}
\end{equation}
where $\mathbf{S} \in \mathbb{R}^{L_s \times D}$, and $L_s$ is the number of patches. Then, these patch tokens are enhanced by a multi-head self-attention (MHA) block. This step allows each patch token to gather information from all other tokens, creating a context-aware representation that captures global temporal dependencies.
\begin{equation}
\mathbf{S}_{\text{attn}} = \mathbf{S} + \text{MHA}(\text{Norm}(\mathbf{S})) \text{.}
\label{eq:mha}
\end{equation}
The resulting sequence, $\mathbf{S}_{\text{attn}}$, represents a set of tokens enhanced at data and model levels, ready for efficient processing. 

\subsection{Attention-Based Patch Sparsification}
\label{sec:}
The next stage incorporates an efficient  mechanism to address the trade-off between accuracy and computational cost. Instead of processing all $L_s$ tokens, the most discriminative ones are dynamically identified and preserved, while the remainder is aggregated. This process begins by computing a ``salience score'' for each enhanced patch token in $\mathbf{S}_{\text{attn}}$. A simple linear projection followed by a Sigmoid activation is employed to learn these scores, $\mathbf{a}_{\text{patch}} \in \mathbb{R}^{L_s}$:
\begin{equation}
\mathbf{a}_{\text{patch}} = \text{Sigmoid}(\text{MLP}(\mathbf{S}_{\text{attn}})) \text{.}
\label{eq:attn_score}
\end{equation}

Given a predefined sparsity ratio $r$, the proposed method identifies the indices of the top-scoring tokens, $\mathcal{I}_{\text{keep}}$. The key feature of the ``soft-sparse'' approach is that the remaining tokens are not discarded. Instead, the less salient tokens (indexed by $\mathcal{I}_{\text{discard}}$) are aggregated via sum pooling into a single representative vector, $\mathbf{h}_{\text{agg}}$:
\begin{equation}
\mathcal{I}_{\text{keep}} = \text{TopK}(\mathbf{a}_{\text{patch}}, \lceil r \cdot L_s \rceil) \text{,}
\label{eq:topk}
\end{equation}
\begin{equation}
\mathbf{h}_{\text{agg}} = \sum_{i \in \mathcal{I}_{\text{discard}}} \mathbf{S}_{\text{attn}, i} \text{.}
\label{eq:agg}
\end{equation}
The final sparsified sequence, $\mathbf{S}_{\text{sparsified}}$, is formed by concatenating the most important tokens with this aggregated summary token.
\begin{equation}
\mathbf{S}_{\text{sparsified}} = [\mathbf{S}_{\mathcal{I}_{\text{keep}}}, \mathbf{h}_{\text{agg}}] \text{.}
\label{eq:sparse}
\end{equation}

This methodology significantly curtails sequence length for subsequent layers. It operates by preserving not only the key discriminative information but also the implicit context held within less salient patches. Consequently, this design ensures that exceptional performance is maintained alongside a dramatic increase in computational efficiency.

\subsection{Dual-Path MoE-Inception Fusion}
\label{sec:fusion}
Once the tokens are sparsified, the model faces a dual challenge: it must simultaneously model the highly specific, individual variations of each subject while also capturing the broad, global temporal patterns shared across the data.

To address this challenge, the proposed method employs a dual-path fusion architecture that processes the $\mathbf{S}_{\text{sparsified}}$ tokens. This architecture splits the task between two specialized, parallel pathways, with each pathway tailored for a specific component of the task. 

\textbf{The MoE Path: Handling Subject-Specific Variations.}
The first path confronts the high inter-patient variability common in medical data. It uses a sparse Mixture of Experts (MoE) network, which maintains a set of $E$ specialized ``expert'' networks (MLPs). Think of these as specialists. For each token, a dynamic gating mechanism $\mathbf{g}_{\text{norm}}$ acts as a dispatcher, selecting the most relevant experts to handle that specific piece of information. This allows the model to adapt its processing to the unique, localized patterns of each subject:
\begin{equation}
\mathbf{H}_{\text{moe}} = \text{MoE}(\text{Norm}(\mathbf{S}_{\text{sparsified}})) \text{.}
\label{eq:moe}
\end{equation}

\textbf{The Inception Path: Capturing Global Patterns.}
While the MoE path excels at this specialized adaptation, it is so focused on localized patterns that it might overlook larger-scale temporal dependencies. To compensate, the second path runs in parallel, designed specifically to capture the global picture. It uses a multi-scale Inception module.
This module applies convolutional filters of different kernel sizes ($k_i$) and a max-pooling layer simultaneously, which is akin to looking at the data through multiple ``lenses'' at once.
This process allows it to extract patterns at various temporal granularities, ensuring the model grasps the overarching temporal structure that the MoE path might miss:
\begin{equation}
\mathbf{H}_{\text{ms}} = \text{Inception}(\text{Norm}(\mathbf{S}_{\text{sparsified}})) \text{.}
\label{eq:inception}
\end{equation}
The outputs from these two complementary paths are then fused with the original input tokens $\mathbf{S}_{\text{sparsified}}$ via residual connections:
\begin{equation}
\mathbf{H}_{\text{combined}} = \mathbf{S}_{\text{sparsified}} + \mathbf{H}_{\text{moe}} + \mathbf{H}_{\text{ms}} \text{.}
\label{eq:fusion}
\end{equation}
This fusion yields a comprehensive representation, $\mathbf{H}_{\text{combined}}$, that is both locally adaptive and globally aware, which is then fed into the final classification head.

\section{Experiments}
This section presents the experimental validation of the proposed DE3S framework. The discussion is structured as follows: First, the experimental setup, datasets, and baselines are detailed. Second, a performance comparison against state-of-the-art methods is presented. Finally, an ablation study analyzes the contribution of each core component.

\subsection{Experimental Setup}
The proposed DE3S model is implemented in PyTorch and trained using the AdamW optimizer. All experiments were conducted on an NVIDIA RTX 4090 graphics processor running Ubuntu 20.04 LTS. 

\subsubsection{Datasets}
The performance of the proposed framework is evaluated on six diverse, publicly available medical time series datasets. These datasets are grouped into two categories based on their clinical application: short-term early prediction and subject-consistency classification. A summary of their statistical details is provided in TABLE~\ref{tab:dataset_stats}.

\textbf{Early Prediction Datasets.} This category includes two sepsis datasets (Sepsis1 and Sepsis2) and a Sickle Cell Clinical (SCC) dataset \cite{shah2019sickle}.

\textbf{Subject-Consistency Classification Datasets.} Subject-consistency evaluation ensures recordings from the same subject remain within the same data split \cite{Kwon2020}, crucial for testing model robustness. This paradigm evaluates the ability to capture subject-specific variations while maintaining generalization. This category includes APAVA and ADFTD (EEG datasets for Alzheimer's and frontotemporal dementia) and PTB (ECG dataset for myocardial infarction). These data are sourced from the Medformer library \cite{wang2024medformer}.

\begin{table}[t!]
\centering
\caption{Statistical Summary of Datasets.}
\label{tab:dataset_stats}
\begin{tabular}{lccccc}
\toprule 
Dataset & \# Samples & \# Train & \# Test & \# Timesteps & \# Features \\
\midrule 
ADFD    & 10,448     & 8,526    & 1,922   & 256          & 19           \\
APAVA   & 1,953      & 1,413    & 540     & 256          & 16           \\
PTB     & 2,647      & 2,159    & 488     & 300          & 15           \\
SCC     & 427        & 341      & 86      & [10, 20]     & 3            \\
Sepsis1 & 472        & 376      & 96      & [2, 24]      & 35           \\
Sepsis2 & 533        & 425      & 108     & [2, 24]      & 35           \\
\bottomrule 
\end{tabular}
\end{table}

\subsubsection{Baselines}
To ensure a comprehensive evaluation, the proposed framework is compared against two main categories of baseline methods:

\begin{itemize}
    \item \textbf{Traditional Machine Learning Methods:} We selected widely-used traditional baselines including XGBoost \cite{XGBoost2016}, RandomForest \cite{Breiman2001}, and OneStepKNN \cite{2023onestepknn}.
    \item \textbf{Advanced Deep Learning Methods:} We included recent state-of-the-art deep learning models designed for time series, namely MPTSNet \cite{mu2025mptsnet}, TimeMixer \cite{wang2023timemixer}, and Medformer \cite{wang2024medformer}.
\end{itemize}

\subsection{Prediction Performance Comparison}
\subsubsection{Early Prediction Results}
\begin{figure}[t]
\centering
\includegraphics[width=\columnwidth]{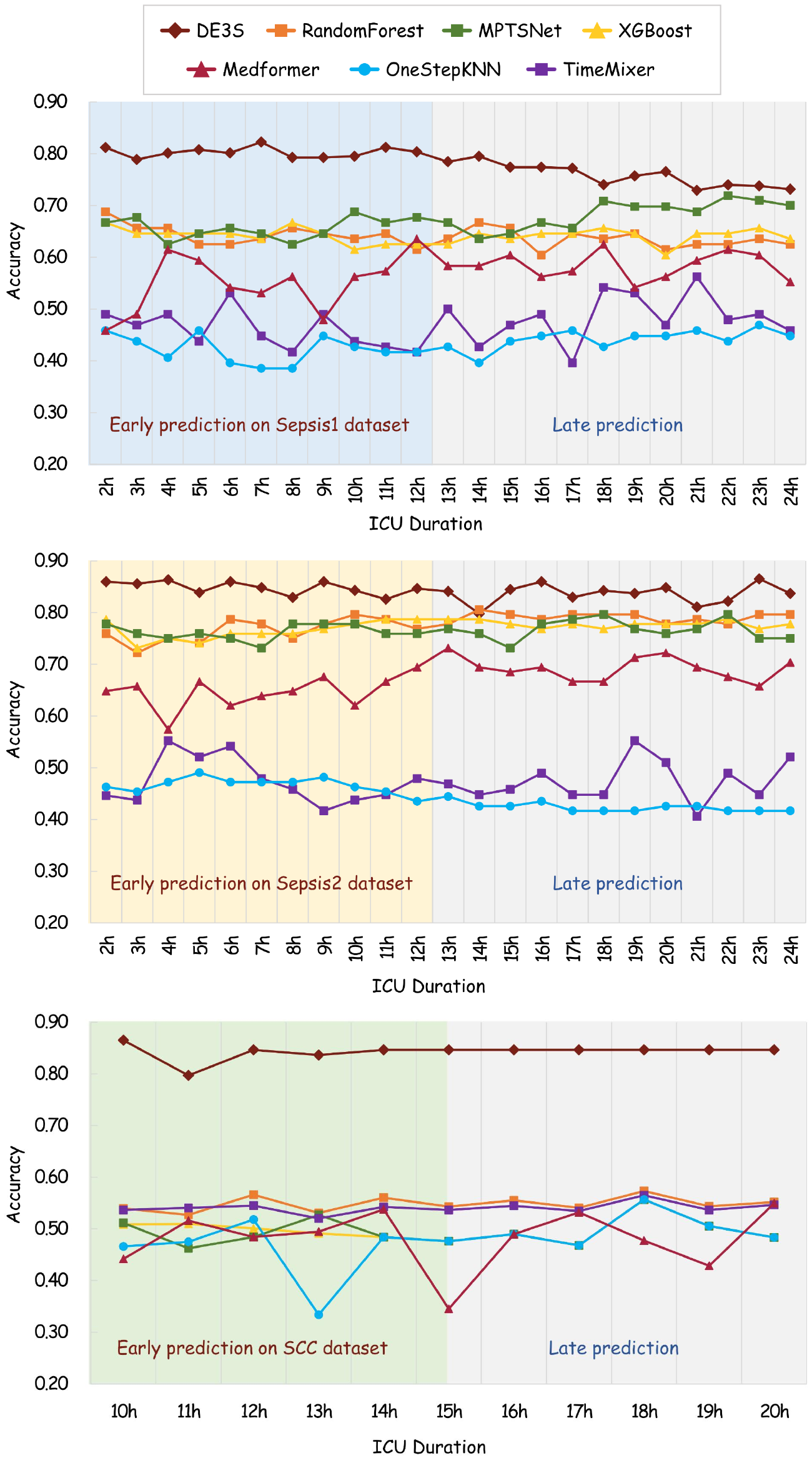}  
\caption{Performance comparison (Accuracy) on the Sepsis1, Sepsis2, and SCC datasets across different temporal lengths. The x-axis represents the observation window, demonstrating the models' early detection capabilities.}
\label{fig:comparison}
\end{figure}

Fig.~\ref{fig:comparison} presents the performance comparison on early prediction datasets. The results show that DE3S achieves consistently superior performance across all three datasets and all observed temporal lengths. The advantage of DE3S is prominent in the earliest prediction windows (e.g., 6-12 hours for sepsis), which are the most critical for timely medical intervention. This robust early detection capability validates DE3S's effectiveness for time-sensitive applications and demonstrates its strong generalization across different medical conditions.

\subsubsection{Subject-Consistency Prediction Results}
TABLE~\ref{comparison exp} shows the performance on subject-consistency datasets, which evaluates the model's ability to capture individual-specific patterns. As demonstrated in TABLE~\ref{comparison exp}, DE3S exhibits superior performance, achieving a top-two rank in 6 out of 9 evaluation metrics (including four first-place finishes). It surpasses the next-best model, Medformer, which notably excelled on the APAVA metric. This comprehensive performance validates DE3S's capacity to handle subject-specific variations while maintaining robust generalization. In comparison, traditional machine learning approaches show clear limitations in capturing the data's temporal complexity and inter-subject variability. Although advanced deep learning models (MPTSNet, TimeMixer, Medformer) achieve better results, their performance still reveals specific gaps that DE3S successfully addresses. Ultimately, the strong overall performance of DE3S across both early prediction and subject-consistency tasks underscores its generalizability and potential clinical utility.

\begin{table}[t!]
\centering
\caption{Performance comparison on subject-consistency datasets. \colorbox{bestblue}{\textbf{Best}} and \colorbox{secondblue}{\textbf{second-best}} results are highlighted by cell background.}
\label{comparison exp}
\begin{tabular}{lcccc}
\toprule 
Dataset & Method & Accuracy & Precision & F1-Score \\
\midrule 
\multirow{7}{*}{PTB} 
 & XGBoost & 0.4465 & 0.4656 & 0.4378 \\
 & RandomForest & 0.5645 & 0.5692 & 0.5532 \\
 & OneStepKNN & 0.5237 & 0.4228 & 0.4586 \\
 & MPTSNet & \cellcolor{secondblue}0.6048 & \cellcolor{secondblue}0.6361 & \cellcolor{secondblue}0.5982 \\
 & TimeMixer & 0.4686 & 0.5445 & 0.4892 \\
 & Medformer & 0.5138 & 0.6208 & 0.5218 \\
 & \textbf{DE3S}(ours) & \cellcolor{bestblue}0.6122 & \cellcolor{bestblue}0.6713 & \cellcolor{bestblue}0.6068 \\
\cmidrule(lr){1-5} 
\multirow{7}{*}{APAVA} 
 & XGBoost & \cellcolor{secondblue}0.7116 & 0.7842 & 0.7229 \\
 & RandomForest & 0.4796 & 0.7458 & 0.4885 \\
 & OneStepKNN & 0.1915 & 0.4330 & 0.1758 \\
 & MPTSNet & 0.6238 & 0.7670 & 0.5972 \\
 & TimeMixer & 0.5933 & 0.7271 & 0.6353 \\
 & Medformer & \cellcolor{bestblue}0.7666 & \cellcolor{bestblue}0.8549 & \cellcolor{bestblue}0.7909 \\
 & \textbf{DE3S}(ours) & 0.6456 & \cellcolor{secondblue}0.8459 & \cellcolor{secondblue}0.7318 \\
\cmidrule(lr){1-5} 
\multirow{7}{*}{ADFD} 
 & XGBoost & 0.4186 & \cellcolor{bestblue}0.3895 & \cellcolor{bestblue}0.3787 \\
 & RandomForest & 0.3523 & \cellcolor{secondblue}0.3732 & \cellcolor{secondblue}0.3273 \\
 & OneStepKNN & 0.1582 & 0.1756 & 0.1554 \\
 & MPTSNet & \cellcolor{secondblue}0.4228 & 0.1803 & 0.2516 \\
 & TimeMixer & 0.4211 & 0.1787 & 0.2506 \\
 & Medformer & 0.4208 & 0.1787 & 0.2505 \\
 & \textbf{DE3S}(ours) & \cellcolor{bestblue}0.4244 & 0.3033 & 0.2728 \\
\bottomrule 
\end{tabular}
\end{table}


\subsection{Ablation Study}
\begin{table}[t!]
\centering
\caption{Ablation study on the Sepsis2 dataset. \colorbox{bestblue}{\textbf{Best}} and \colorbox{secondblue}{\textbf{second-best}} results are highlighted by cell background.}
\label{tab:ablation}
\begin{tabular}{lccc}
\toprule
Variants of DE3S & Accuracy & Precision & F1-Score \\
\midrule
\textbf{DE3S}(full) & \cellcolor{bestblue}0.8971 & \cellcolor{bestblue}0.8996 & \cellcolor{bestblue}0.8966 \\
w/o Data Enhancement & \cellcolor{secondblue}0.8785 & \cellcolor{secondblue}0.8843 & \cellcolor{secondblue}0.8791 \\
w/o Model Enhancement & 0.8715 & 0.8728 & 0.8708 \\
w/o Dual Enhancement & 0.8489 & 0.8699 & 0.8481 \\
w/o Patch Sparsification & 0.8651 & 0.8563 & 0.8641 \\
w/o Inception & 0.8483 & 0.8429 & 0.8471 \\
w/o MoE\&Inception & 0.8021 & 0.8003 & 0.8036 \\
\bottomrule
\end{tabular}
\end{table}

To validate the design rationale, we conduct systematic ablation studies on the Sepsis2 dataset to quantify each component's contribution to addressing the core challenges of weak early-stage signals and subject-specific variations. TABLE \ref{tab:ablation} highlights the hierarchical importance of our proposed components. The Dual Enhancement mechanism proves essential for addressing weak early signals. Notably, removing model-level enhancement (accuracy drops to 0.8715, -2.56\%) causes greater performance degradation than removing data-level enhancement (0.8785, -1.86\%), validating that attention-based model-level enhancement plays a more critical role in capturing global context from sparse observations. The Soft Shape Token Sparsification mechanism demonstrates effective balance, successfully reducing computational complexity via attention-guided aggregation while preserving discriminative power (0.8651 accuracy). Most importantly, the Dual-Path MoE-Inception Fusion shows the largest contribution, as its complete removal incurs the most severe performance degradation (0.8021 accuracy, -9.5\%). This validates our core architectural hypothesis: synergistic modeling of both local subject-specific variations and global temporal dependencies is indispensable for effective early medical prediction. 

\section{Related Work}
This section reviews key relevant works: early time series classification and shapelet-based methods.

\subsection{Early Time Series Classification}

Early Time Series Classification (ETSC) is critical in time-sensitive domains like healthcare~\cite{ismail2019deep} and biomedcine~\cite{zhao2023selecting}. Traditional machine learning methods such as OneStepKNN~\cite{2023onestepknn}, XGBoost~\cite{XGBoost2016}, and RandomForest~\cite{Breiman2001} rely on statistical learning but struggle to amplify weak early signals in short-term medical time series, due to the noise and class imbalance~\cite{gao2025data, luo2024efficient} issues. Deep learning approaches such as MPTSNet~\cite{mu2025mptsnet}, with its multiscale periodic feature integration, Medformer's~\cite{wang2024medformer} multi-granularity patching, and TimeMixer's~\cite{wang2023timemixer} hybrid temporal modeling, advance automatic feature extraction but remain limited in capturing faint early medical signals. They focus on periodicity, granularity, or hybrid dynamics may overlook transient biomarkers or dilute sparse critical features. To address this, our DE3S method proposes a dual-enhancement strategy specifically designed to amplify these weak early signals for more robust classification.

\subsection{Shapelet-Based Methods}
Shapelets~\cite{ye2009time}, discriminative subsequences representing class characteristics, are foundational in time series analysis. Traditional methods involve costly enumeration and evaluation of all potential candidates~\cite{lines2012shapelet,hills2014classification}, limiting scalability. Subsequent advances focused on learning shapelets during model training, with ``learnable shapelets'' often implemented as convolutional filters~\cite{hou2016efficient}. Recent innovations include TimeCSL's~\cite{TimeCSL2024} unsupervised contrastive learning for general-purpose shapelets enabling interpretable pattern extraction for versatile tasks; RSFCF's~\cite{liu2022RSFCF} efficiency improvements via random shapelet selection, restricted matching ranges, and canonical feature embedding within a random forest framework; and ShapeFormer's~\cite{ShapeFormer2024} integration of class-specific via offline discovery, Shapelet Filter and generic transformer modules to capture multivariate dependencies.
Despite these progressions, computational intensity remains a challenge. This stems from factors such as contrastive learning overhead, ensemble complexity, or transformer architecture demands, alongside inefficient handling of vast potential shapelet spaces.
Drawing inspiration from foundational work on time series shapelets~\cite{Liu2024_shape}, which identify critical local sub-sequences, the proposed DE3S advances this field. However, rather than extracting traditional, rigid shapelets, DE3S introduces a novel soft-sparse shape learning mechanism designed to learn the shape features of time-series patches. This mechanism employs an attention-based strategy to dynamically aggregate informative patterns, preserving critical information while enhancing computational efficiency.

\section{Concluding Remarks}
This paper proposes DE3S, a novel framework designed to address fundamental challenges in early medical time series classification, such as weak signals and subject-specific variations. More specifically, the DE3S architecture introduces three coordinated innovations: 1) The Dual Enhancement Mechanism robustly amplifies critical signals at both the data and model levels; 2) Attention-based Patch Sparsification efficiently identifies and preserves discriminative patterns while discarding redundant computations; and 3) The Dual-Path MoE-Inception Fusion provides a novel mechanism to simultaneously model both subject-specific dynamics and global temporal patterns. Experimental validation across six medical datasets confirms that DE3S achieves consistent state-of-the-art performance, demonstrating that domain-specific, robust models addressing both local and global context can outperform generic solutions. For future research, exploring multi-modal extensions, developing medical-tailored explainability techniques, and utilizing the proposed method to boost prospective clinical trials are promising.

\bibliographystyle{IEEEtran}
\bibliography{references}

\end{document}